\documentclass[letterpaper, 10 pt, conference]{ieeeconf}  

\IEEEoverridecommandlockouts                              

\usepackage{graphicx}
\usepackage{subcaption}
\usepackage{booktabs} 
\usepackage{tabulary}
\usepackage{setspace}
\usepackage{bm}
\usepackage{amsthm,amsmath,amssymb} 
\usepackage{color}
\usepackage{cite}
\usepackage{multicol}
\usepackage{algorithm}
\usepackage{algorithmic}
\usepackage{tabularx}
\usepackage{multirow}
\usepackage{array}
\usepackage{soul}
\usepackage{upgreek}
\usepackage{mathtools}

\title{\LARGE \bf Optimal Control-Based Baseline for Guided Exploration in Policy Gradient Method}

\author{Xubo Lyu$^{1}$, Site Li$^{1}$, Seth Siriya$^{2}$, Ye Pu$^{2}$ and Mo Chen$^{1}$ 
\thanks{$^{1}$School of Computing Science, Simon Fraser University, BC, Canada {\tt\scriptsize \{xlv, sitel\}@sfu.ca,  mochen@cs.sfu.ca}; $^{2}$Department of Electrical and Electronic Engineering, University of Melbourne, Vic, Australia
        {\tt\scriptsize ssiriya@student.unimelb.edu.au, ye.pu@unimelb.edu.au}}%
}

\begin{document}

\maketitle
\thispagestyle{empty}
\pagestyle{empty}

\begin{abstract}
In this paper, a novel optimal control-based baseline function is presented for the policy gradient method in deep reinforcement learning (RL). The baseline is obtained by computing the value function of an optimal control problem, which is formed to be closely associated with the RL task. In contrast to the traditional baseline aimed at variance reduction of policy gradient estimates, our work utilizes the optimal control value function to introduce a novel aspect to the role of baseline -- providing guided exploration during policy learning. This aspect is less discussed in prior works. We validate our baseline on robot learning tasks, showing its effectiveness in guided exploration, particularly in sparse reward environments.

\end{abstract}

\section{Introduction}
\label{sec:intro}
Deep reinforcement learning has achieved remarkable success in robotics \cite{schulman2015trust, schulman2017ppo, Lillicrap2016ContinuousCW, johannink2019residual}. 
One of the key techniques behind the success is the policy gradient method. It uses gradient descent to directly optimize a parameterized control policy with sampled task data, enabling effective learning of high-dimensional and continuous policies but yielding gradient estimates with high variance.

The baseline function is a commonly used component in the policy gradient methods to mitigate its high variance of gradient estimates \cite{greensmith2004variance, hofmann2015variance}.
A baseline is typically a state-dependent function subtracted from the observed total reward, also called return, resulting in a shifted return. This shifted return yields an unbiased estimate of the gradient with reduced variance. A baseline can have various choices, including the value-based \cite{schulman2015trust, schulman2017ppo, gu2016q}, gradient norm-based \cite{peters2008reinforcement}, and trajectory-based forms \cite{cheng2020trajectory}.

Previous research on baseline \cite{gu2016q, peters2008reinforcement, liu2017action,wu2018variance,cheng2020trajectory}
has primarily focused on mitigating gradient variance.
While reducing variance is essential, it may not be adequate to ensure policy success, given the challenges of insufficient guidance and exploration in sparse feedback environments. Although other methods exist to tackle these challenges, there is limited research that addresses them specifically from the perspective of the baseline function. Therefore, we aim to investigate a novel aspect of baseline, particularly its role in addressing the problem of inefficient exploration in RL.

In this paper, we introduce a novel, optimal control-based baseline to provide guided exploration for policy gradient methods. 
The baseline is obtained by solving a value function for an optimal control problem. This problem stems from the original RL task, from where a cost function and a {coarse} mathematical model are identified to approximately describe the objective and the robot system involved in the RL task. This allows the use of optimal control techniques for efficient value computation. Then the value function of the optimal control problem can be used as the baseline for the policy gradient method to solve the RL task. The key insight here is that the optimal control value function can offer essential prior information related to the RL task, such as the robot system dynamics, thus providing guided exploration for policy gradient RL.

We empirically evaluate our method using the standard policy gradient algorithm on a variety of robot learning tasks and provide a thorough analysis of the baseline's effect on guided exploration, particularly under sparse reward setups.

\section{Related Work}
\subsection{Baselines in Policy Gradient RL}
Various baselines have been proposed in policy gradient RL to mitigate its high variance of gradient estimate. A foundational study \cite{greensmith2004variance} established the theoretical bounds of estimated gradient variance induced by commonly used baselines such as the value function. To further lower the variance, recent studies have expanded the range of baseline forms, such as separate baselines for every coefficient of the gradient \cite{peters2008reinforcement}, Taylor expansion of the off-policy critic as baseline \cite{gu2016q}, state-action dependent baselines \cite{liu2017action, wachter2006implementation}, and a trajectory correlation-based baseline \cite{cheng2020trajectory}. However, most baselines are designed for variance reduction while the potential role of baseline in other aspects, such as providing guided exploration for policy learning, is less studied.

\subsection{Robot System Models from Control Theory}
Classical control theory provides mathematical models for a range of robots based on their physical dynamics. For example, the Dubins car model \cite{chen2015safe, chen2016multi, chen2016exact} simplifies the motion of vehicles by assuming constant speed and position-steering. The extended Dubins car model \cite{9683031, lyu2020ttr} allows variable speeds and turning rates,  providing a more realistic representation. Both models are often used to describe differential-wheel robots and four-wheeled vehicles. In addition, the planar quadrotor model \cite{chen2021fastrack, boris2019barc} simplifies the quadrotor motion from 3D to 2D plane and the attitude model focuses solely on 3-axis rotational behaviour \cite{tahir2019state} while point-mass model \cite{romero2022model} only focuses on translation. Various other models, such as those for landing robots \cite{lavalle2006planning}, and pendulum-like systems \cite{pati2014modeling} are available. These models provide simplified abstractions of complex real-world robot models but still capture vital system priors with the potential to enhance modern robot learning methods, such as RL.

\begin{figure*}[!htp]
    \centering    \includegraphics[width=\linewidth]{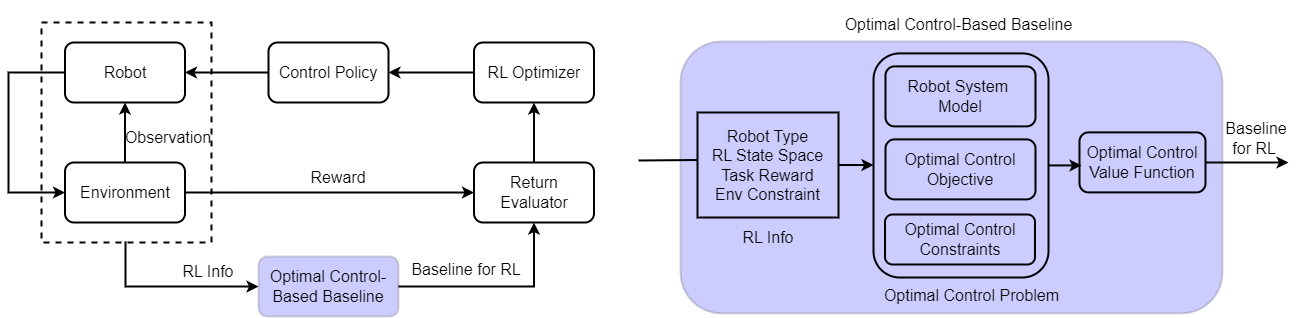}
    \caption{\small Overview of our method. \textbf{Left}: We propose a novel baseline function for policy gradient RL. Our method involves extracting RL info from the robot and environment, which is used to formulate an associated optimal control problem. Subsequently, we compute the optimal control value function and used it as a baseline for the policy gradient RL. \textbf{Right}: The RL info encompasses crucial aspects of the RL problem, including robot types, RL full state, task reward, and more. This RL info serves to form the key components of an optimal control problem, which include the robot system model, objectives, and constraints. Techniques like Model Predictive Control (MPC) can then be employed to compute the value function, which can be utilized as an RL baseline.
    }
    \label{fig:intro}
\end{figure*}

\subsection{Relations to Our Work}
Our work falls into the realm of introducing a novel baseline for policy gradient RL. Unlike prior work which focused on reducing gradient variance, this new baseline prioritizes providing valuable guidance in a subspace of the RL full state space during policy learning. Furthermore, our baseline is formed by utilizing the value function derived from an associated optimal control problem. It innovatively leverages well-developed robot system models from control theory to incorporate essential robot priors to provide guidance exploration for policy gradient RL.

\section{Optimal Control-Based Baseline}
\label{sec:method}

This section describes the process of designing an optimal control-based baseline. It starts by extracting the key RL info from the RL task to form an associated optimal control problem. Then the value function of this optimal control problem is computed and serves as a baseline for policy gradient RL to solve the original RL task, as shown in Fig.~\ref{fig:intro}. 

\subsection{Extraction of RL Info}
A given RL problem often has key information characterizing its core aspects, which is denoted as RL Info in this work. There are in total four common components of RL info: robot type, state space, task reward, and environment constraints. This info can be extracted from the given robot and environment setup. 
\begin{enumerate}
    \item {Robot Type}. Robots can have diverse types, like cars, quadrotors, landing robots, etc.
    \item {RL State Space}. RL's full state often includes the physical state of the robot such as its position and velocity as well as sensory observation such as pixels or LiDAR readings. This state could be high-dimensional. 
    \item {Task Reward}. It is numerical feedback provided to the robot at each time step to evaluate the robot's actions. 
    \item {Environment Constraints}. Typical constraints include factors such as maximum episode length,  the valid range of state and action spaces, and physical limitations such as obstacles in the navigation task.
\end{enumerate}

\subsection{Formulation of Optimal Control Problem}
\label{subsec:formation_oc}
 Once the RL info mentioned above is obtained, an associated optimal control problem can be formed, which consists of three primary components: robot system model, optimal control objective, and related constraints.

\subsubsection*{\textbf{Robot System Model}}
In control theory, there are many known models developed based on the physical dynamics of robot systems. Despite being abstracted and simplified from true dynamics, these models allow efficient optimal control computation and have the potential to be used in RL problems to mitigate its data inefficiency especially when the RL environment model is unknown and hard to learn. Thus, we aim to leverage those control theory models to calculate an optimal control value function, which can be used as a baseline function for policy gradient RL to provide guidance.

\begin{table}[]
\centering
\small
\begin{tabular}{|c|c|}
\hline
Robot Type & Robot System Model \\ \hline
Car-like system & \begin{tabular}[c]{@{}c@{}}3D Dubins car \cite{dubins1957curves} \\ 5D extended Dubins car \cite{9683031} \\ 4D bicycle model \cite{francis2016models} \end{tabular} \\ \hline
Quadrotor & \begin{tabular}[c]{@{}c@{}}6D planar quadrotor \cite{lyu2020ttr} \\ 6D rotation model \cite{tahir2019state} \\
3D point mass model \cite{romero2022model} \end{tabular} \\ \hline
\end{tabular}%
\caption{\small A list of available robot system models. A 3D Dubins car has three states -- position ($x$, $y$) and its heading angle ($\theta$), describing the dynamics of a simple differential-wheeled robot. Its 5D extended version involves speed $v$ and turn rate $\omega$ as additional states. A 4D bicycle model describes the motion of a four-wheeled vehicle with positions, heading and specifically steering angle. For quadrotors, the 6D planar model focuses on 2D plane motion with states of positions, velocities, and 1-axis rotation. The 6D rotation model focuses on the 3-axis rotation without translation while a simple point-mass model focuses solely on 3D translation without any rotational dynamics.}
\label{tab:robot_system_model}
\end{table}

The robot system model is a mathematical equation describing robot physical dynamics and can be represented by a state-space Ordinary Differential Equation (ODE), denoted by $\dot{\mathrm{x}}(t)=g(\mathrm{x}(t),\mathrm{u}(t))$ where $\mathrm{x}(t)$ and $\mathrm{u}(t)$ are the state and control of an optimal control problem at continuous time $t$ and $g(\cdot, \cdot)$ is the model function. 
Finite difference schemes such as the Forward Euler Method can be applied to this ODE to obtain its discrete version $\mathrm{x}_{k+1}=g(\mathrm{x}_k, \mathrm{u}_k)$, where $k$ is the discrete time step.

This model can be heuristically selected from control theory based on RL info. To do that, we first categorize the robot system by its type (e.g., car-like, quadrotor or others) to select a set of candidate models, referring to Table.~\ref{tab:robot_system_model}. Then we assess candidate models with the following two criteria: 
\begin{enumerate}
    \item The components of optimal control state $\mathrm{x} \in \mathbb{X}$ should be a subset from the components of RL full state $s \in \mathbb{S}$, where $\mathbb{X}$ is the subspace of $\mathbb{S}$.
    \item The RL task reward $r(\cdot)$ can be written in terms of the components of $\mathrm{x}$, that is $r(s) \coloneqq r(\mathrm{x})$. This criterion is typically met when the RL task reward is sparse and related to a limited set of states.
\end{enumerate}

\subsubsection*{\textbf{Optimal Control Objective}} The optimal control objective specifies the desired outcome that an optimal control problem aims to achieve while satisfying the robot system model. It is often described in Eq.~\eqref{eq:oc_objective} as the minimization of cumulative cost subject to the system model constraint,
\begin{equation}
\begin{aligned}
&\min_{\mathrm{u}_{0:T-1}}\sum_{k=0}^{T-1} c(\mathrm{x}_k, \mathrm{u}_k) \\
\text{s.t.}\;\; 
\mathrm{x}_{k+1}&=g(\mathrm{x}_k, \mathrm{u}_k) \text { for } k=0, \ldots, T-1
\label{eq:oc_objective}
\end{aligned}
\end{equation}
where $c(\mathrm{x}_k, \mathrm{u}_k)$ is the step cost function defined over optimal control state $\mathrm{x}_k$ and control input $\mathrm{u}_k$ at each discrete time step $k$, and $\mathrm{x}_{k+1} = g(\mathrm{x}_k, \mathrm{u}_k)$ is the discrete-time robot system model. The key to deriving this objective is to find the appropriate cost function. Note that despite the presence of the RL reward, its direct utilization as the cost in Eq.~\eqref{eq:oc_objective} is generally impractical due to the potential difficulty in obtaining an optimal control solution.

The cost function  $c(\mathrm{x}_k, \mathrm{u}_k)$ has multiple choices. One common choice is set-point tracking, where the cost can be defined by penalizing the error between the current and desired state whilst optionally minimizing the control effort 
\begin{equation}
    c(\mathrm{x}_k, \mathrm{u}_k) = \alpha e_{\mathrm{x}}(\mathrm{x}_k, \mathcal{G}) + \beta e_{\mathrm{u}}(\mathrm{u}_k), 
    \label{eq: oc_cost}
\end{equation}
where $e_{\mathrm{x}}$ and $e_{\mathrm{u}}$ are state and control error functions. To use this cost function, we assume the existence of a goal region $\mathcal{G}$ from the original RL problem. For instance, $e_{\mathrm{x}}$  can be expressed as the Euclidean distance between 2D positions in navigation tasks while $e_{\mathrm{u}}$ can be expressed as ${l}^2$-norm of control input $\lVert \mathrm{u}_k \rVert_2$.
The cost function coefficients $\alpha, \beta$ can be flexibly adapted to obtain optimal control solutions.

\subsubsection*{\textbf{Optimal Control Constraints}} This includes additional limitations that must be obeyed when solving an optimal control problem. Typical constraints include the range of state space, the initial state, and the positions of obstacles in a collision-avoidance scenario. Those constraints can be added to the optimal control problem to rectify the solution.

\subsection{Calculation of Optimal Control Value Function}
\label{subsec:training_data_generation}
We use Model Predictive Control (MPC) to solve the formulated optimal control problem and obtain the associated optimal control value function. MPC  \cite{ye2014mpc, ye2018mpc}  is a well-known control technique that repeatedly solves the optimization problem in an iterative manner, and is used as a representative optimization method in this work.

We first uniformly generate a number of initial states from the optimal control state space. With these initial states and objective function Eq.~\eqref{eq:oc_objective}, MPC produces a set of feasible trajectories, denoted by $\{\mathrm{\tau}_j|j=0,1,2,..., N\}$, where $N$ is the number of trajectories and each trajectory has a horizon of length $H_j$. From each trajectory ${\tau}_j$, we select every state $\mathrm{x}_m^j$, which is the $m$th state of the $j$th trajectory and compute its value ${V}^{oc}(\mathrm{x}_m^j)$ following Eq.~\eqref{eq:Bellman_backup}, where $i$ is sum index.
\begin{equation}
\label{eq:Bellman_backup}
\begin{array}{l}
{V}^{oc}\left(\mathrm{x}_{m}^j\right)=\sum_{i=m}^{H_j} \gamma^{i-m} r(\mathrm{x}_m^j)
\end{array}
\end{equation}

This value sums up the discounted reward of each state along the whole trajectory span, where the $\gamma^m$ means the discount factor of $m$th step of the trajectory. Given the states and their corresponding values, we form a discrete dataset $\mathcal{D} = \{(\mathrm{x}_m^j, V^{oc}(\mathrm{x}_m^j))\mid \forall j, m\}$. Then, we can employ any regression model (e.g. neural network) to fit a continuous value function ${V}^{oc}$  based on $\mathcal{D}$ for arbitrary states.

\subsection{Optimal Control Baseline for Policy Gradient RL}
The optimal control value function calculated above can be directly used in the policy gradient RL as the baseline function. Particularly, we consider a generic form of policy gradient RL that involves Generalized Advantage Estimation \cite{Schulman2016HighDimensionalCC} given by Eq.~\eqref{eq:gae actor critic}. $G_{t}^{\lambda}$ is the TD$(\lambda)$ return \cite{sutton1988learning} considering the weighted average of $n$-step returns for $n=1,2,...\infty$ via parameter $\lambda$. $\pi_{\theta}$ is the RL parameterized policy that selecting action $a_t$ based on the state $s_t$ at each time step $t$. 
The baseline function, denoted as $b(s_t)$, can take various forms, such as the widely used on-policy value function.
\begin{equation}
\small
\label{eq:gae actor critic}
    \nabla_{\theta} J\left(\pi_{\theta}\right)=\underset{\tau \sim \pi_{\theta}}{\mathbf{E}}\left[\sum_{t=0}^{T} \nabla_{\theta} \log \pi_{\theta}\left(a_{t} \mid s_{t}\right)\left(G_{t}^{\lambda} - b(s_t)\right)\right]
\end{equation} 

In this work, we simply use the optimal control value function to be a novel form of baseline function: $b(s_t) = V^{oc}(\mathrm{x}_t)$, where $\mathrm{x}_t$ is extracted from $s_t$. We integrate it into Eq.~\eqref{eq:gae actor critic} to estimate policy gradient $\nabla_{\theta} J\left(\pi_{\theta}\right)$ for optimizing policy $\pi_{\theta}$ to solve the robot learning task. This optimal control-based baseline can address challenging robot learning tasks with limited reward feedback, ensuring sufficient exploration.

\section{Results}
\label{sec:result}
In this section, we present results on the performance of the proposed method and name our method as ``OC baseline''.
\subsection{Demo Example: Differential-Drive Car Navigation}
\begin{figure}[htbp]
    \centering
	\begin{minipage}{\linewidth}
     \centering
 \includegraphics[width=0.7\textwidth]{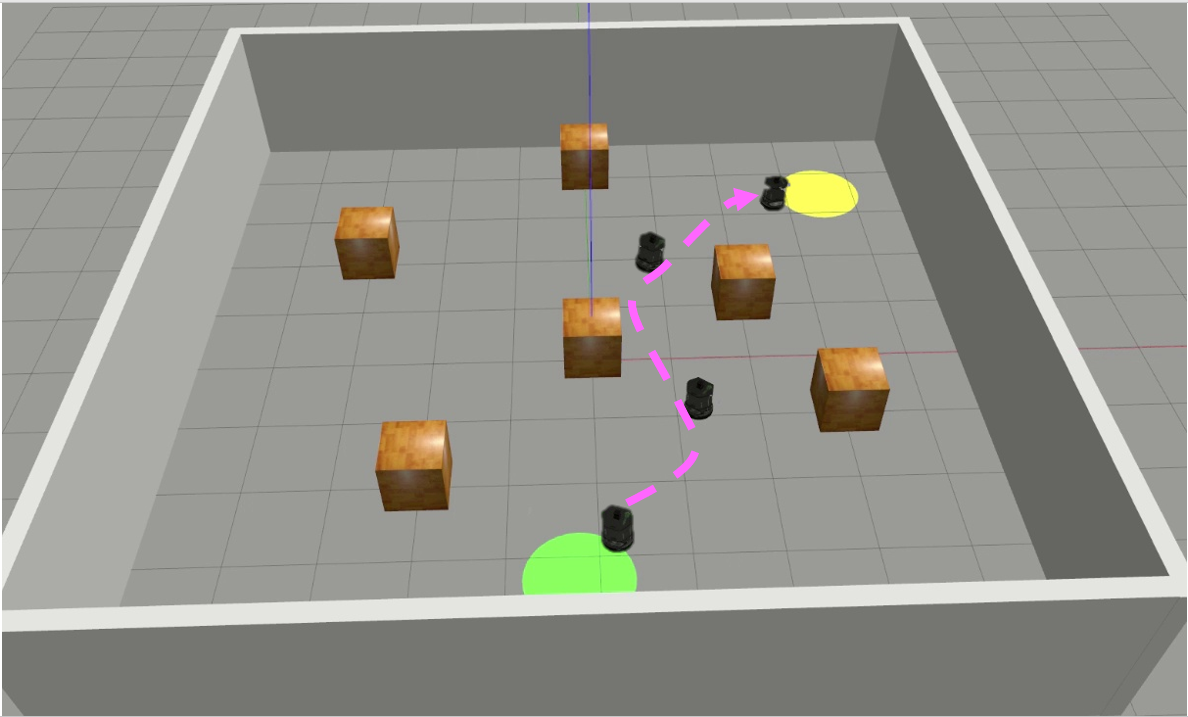}
		\caption{\small Car navigation environment}
		\label{fig:car_env}
	\end{minipage}
\end{figure}

We first evaluate our method on a simulated TurtleBot \cite{amsters2019turtlebot},
which resembles a differential drive car 
navigating in an unknown environment with multiple static obstacles. The car aims to learn a collision-free control policy via trial and error to move from a starting area to a goal area. We use the Gazebo simulator \cite{Koenig-2004-394}
to build the robotic system and environment for the target RL problem, as shown in Fig.~\ref{fig:car_env}.

Following the steps described in the methodology, the original RL problem provides info to form the optimal control baseline: (1) robot type is a car, (2) RL full state is $s = (x, y, \theta, v, \omega, d_1, ...,d_8)$, which are 2D position, heading angle, current speed, current turning rate, and eight-dimensional lidar sensory data. (3) RL task reward is shown in Eq.~\eqref{eq:rew general}, where $\mathcal{G}$ refers to the goal area, $\mathcal{C}$ the collision area and $\mathcal{I}$ intermediate area involving neither collision nor goal. 
All these areas are with regard to 2D position $p = [x,y]^{\top}$.
\begin{equation}
\small
\label{eq:rew general}
    r(s)=
    \begin{cases}
    0&\;(x,y)\in\mathcal{I}\\
    +1000& \;(x,y)\in\mathcal{G}\\
    -400& \;(x,y)\in\mathcal{C}
    \end{cases}
\end{equation}

Environment constraints include start area $\Gamma$, goal area $\mathcal{G}$, obstacle area $\mathcal{O}$, and map size. According to the robot type and key state variables of RL reward, we choose the 3D Dubins car model \cite{dubins1957curves} as an abstraction of the true robot as it captures the 2D positional states and the simplest car motion which is computationally efficient. Its ODE is defined as:
$$
\dot{\mathrm{x}}=g(\mathrm{x}, \mathrm{u})=\left[\begin{array}{c}
\dot{x} \\
\dot{y} \\
\dot{\theta}
\end{array}\right]=\left[\begin{array}{c}
v \cos (\theta) \\
v \sin (\theta) \\
\omega
\end{array}\right],
$$
where state $\mathrm{x}=[x, y, \theta]^{\top}$ and control input $\mathrm{u}=[\omega]^{\top}$, $v$ is the constant speed. 

Then we construct the objective function according to Eq.~\eqref{eq:oc_objective}. As we discussed before, the key step here is designing a proper cost function. In the optimal control problem, the cost function should be designed based on the goal of the related RL problem and its reward function. In this task, the goal is to navigate the car to the goal area, so a typical cost function can be the 2D distance between the robot and the goal. As the distance to the goal decreases, the cost becomes smaller, otherwise, the cost becomes larger. A commonly used mathematical expression of such cost function can be in the quadratic form, written as follows, where $p_k$ is the 2D position at step $k$ and $p^*$ is the goal position, $Q$ and $R$ are coefficients of state and control to describe their relative importance weights.
\begin{equation}
\small
\begin{aligned}
\min_{\mathrm{u}_{0:H-1}}\sum_{k=0}^{H-1}&\left[\left(p_k-p^*\right)^{\top} \gamma^{k+1} Q\left(p_k-p^*\right)+{\omega}_k^{\top} R{\omega}_k\right]\\
\text { s.t. } \quad & {\mathrm{x}_{k+1}}=g(\mathrm{x}_k,\mathrm{u}_k)\\
& \text{initial state } \mathrm{x}_0 \in \Gamma\\
& \text{final state } \mathrm{x}_H \in \mathcal{G} \\
& p_k \notin \mathcal{O}\\
& {\forall} p_k \in \text{map area}.
\end{aligned}
\label{eq: mpc for car}
\end{equation}
The value function can be calculated based on Eq.~\eqref{eq: mpc for car}. Particularly, we used the IPOPT MPC solver \cite{W_chter_2005} to get a set of feasible trajectories and associated state-value dataset $\mathcal{D}=\{(\mathrm{x}_i, V^{oc}(\mathrm{x}_i)) \mid i = 0,1,2,..., N\}$ following Eq.~\eqref{eq:Bellman_backup}. We then use a simple neural network as a regression model to fit a continuous value function $V^{oc}(\mathrm{x})$. Finally, we utilize $V^{oc}(\mathrm{x})$ as the baseline function of a typical policy gradient algorithm, Proximal Policy Optimization (PPO) \cite{schulman2017ppo}, following Eq.~\eqref{eq:gae actor critic} to address the learning-based navigation task.

\begin{figure}[htbp]
    \centering
        \begin{minipage}{\linewidth}
        \centering
        \includegraphics[width=\linewidth]{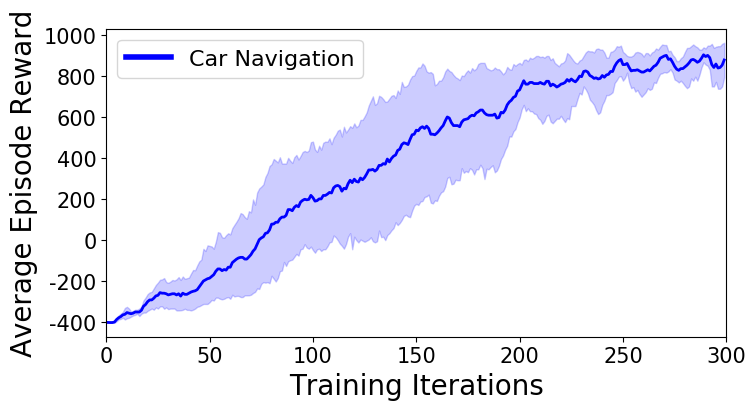}
        \caption{The car navigation reward performance}
        \label{fig:car_nav}
        \end{minipage}
\end{figure}
    As depicted in Fig.~\ref{fig:car_env}, our method effectively learns a collision-free, goal-reaching policy. The policy is able to navigate cautiously through open spaces, skillfully avoid obstacles, and ultimately reach the desired goal along the trajectory in pink. Fig.~\ref{fig:car_nav} shows the car navigation training results that are summarized across 10 different trials. The curves show the mean reward and the shaded area represents the standard deviation of all trials. As depicted in the figure, as training iterations progress, the cumulative reward steadily increases and continues to rise until a gradual convergence towards a stable value. At 300 iterations the reward stabilizes around 800, which implies that the agent may have learned a relatively effective and stable policy.
\subsection{Analysis of Guided Exploration}

\begin{figure}[htbp]
    \centering
    \includegraphics[width=0.9\linewidth]{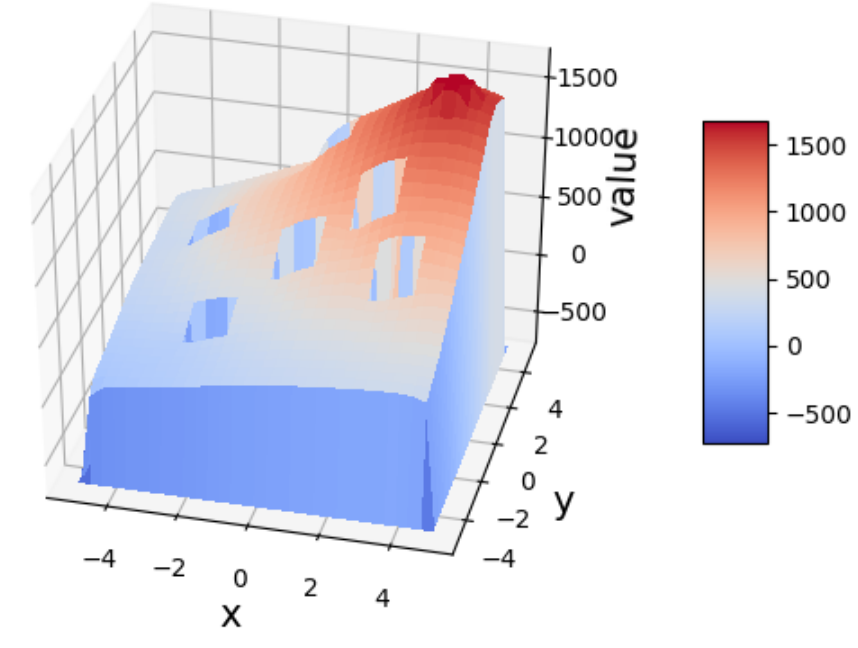}
    \caption{The optimal control value heatmap for car navigation}
    \label{fig:car_heatmap}
\end{figure}

We conduct analysis on the primary benefit of our method in offering guided exploration for policy gradient RL. We illustrate a 2D plane value heatmap which is calculated from the Dubins Car optimal control problem. From Fig.~\ref{fig:car_heatmap}, we can identify value distribution from any initial state to the goal. Regions with darker red colours correspond to higher values. The square concaves correspond to obstacles, denoting significantly lower values as they are to be avoided. With this value function as an RL baseline, it captures the state value priors related to the collision-free, goal-reaching areas in the 2D subspace of the RL full state, thereby providing valuable guidance for policy learning.

\begin{figure}[htbp]
    \centering
    \includegraphics[width=\linewidth]{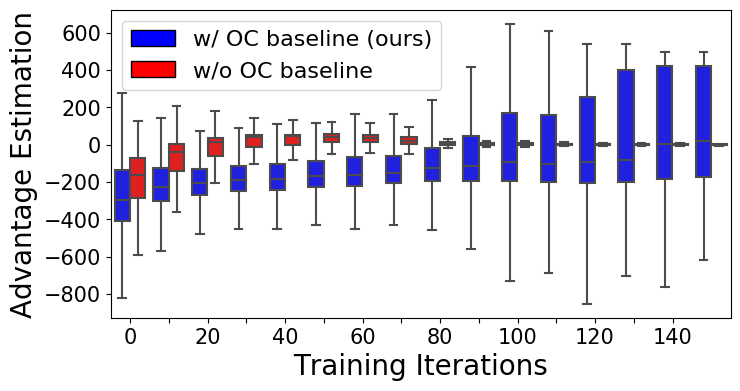}
    \caption{Policy advantage estimation w/ and w/o OC baseline.}
    \label{fig:adv_car}
\end{figure}

We also investigate the cause of exploration by analyzing the advantage estimation in the car example. The advantage $A^{\pi}(s,a)$ is a function that drives the policy update in policy gradient RL by measuring how much better an action $a$'s return than the baseline and is defined by $A^{\pi}(s_t,a_t) = G_{t}^{\lambda} - b(s_t)$ in Eq.~\eqref{eq:gae actor critic}.
In Fig.~\ref{fig:adv_car}, we depict advantage estimation with and without the OC baseline (i.e. on-policy value function $V^{\pi}$ as baseline), averaging every 10 iterations. With the OC baseline, policy gradient RL maintains a wider range of advantage estimation during the early learning stage. Conversely, without the OC baseline, the advantage estimation gradually narrows around zero. This indicates that the OC baseline consistently drives the exploration of diverse actions when initial actions are unfavourable ($A \leq 0$), while without OC baseline, the policy may get stuck in undesired local optima due to negligible advantage estimates, as $V^{\pi}(s_t)$ will be gradually fitted towards $ G_{t}^{\lambda}$.

\subsection{Compare with Other methods}
\begin{figure}[htbp]
    \centering
    \includegraphics[width=\linewidth]{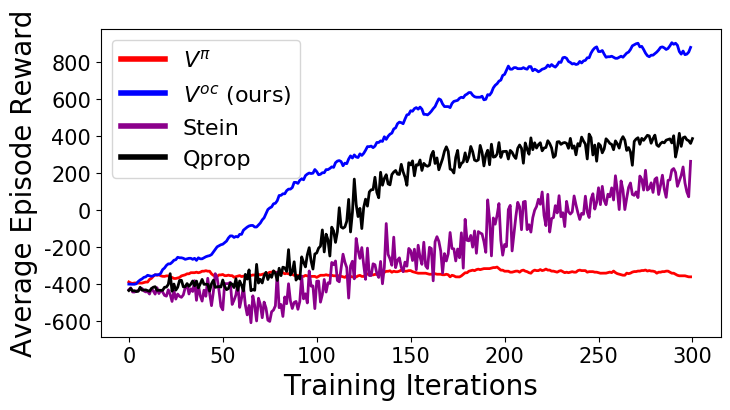}
    \caption{Comparison with other forms of baseline function.}
    \label{fig:comp}
\end{figure}

We compare with existing methods on the car navigation problem. Particularly, we select three other baseline functions and denote them by $V^{\pi}$ \cite{schulman2017ppo}, Stein \cite{liu2017action} and Qprop \cite{gu2016q}. $V^{\pi}$ is the on-policy value function and is widely used as the baseline for actor-critic RL algorithms. Stein refers to a state-action dependent baseline based on the Stein Identity while Qprop refers to a baseline derived from a Taylor expansion of the off-policy critic. The comparative analysis, as depicted in Fig.~\ref{fig:comp}, highlights the superior performance of our OC baseline, particularly in terms of achieving the highest average episodic reward. Our findings suggest that while baseline functions aimed at variance reduction are effective in many scenarios, they may not always provide feasible control solutions, especially when dealing with tasks with sparse rewards and requiring more guidance.
\subsection{Ablation Study}
\begin{figure}[htbp]
    \centering
    \includegraphics[width=\linewidth]{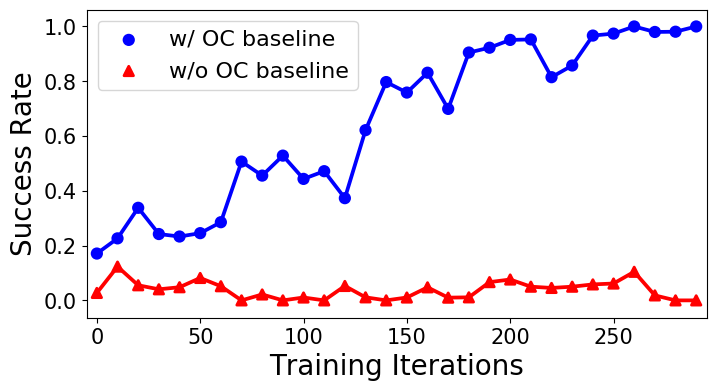}
    \caption{Ablation study of our OC baseline method}
    \label{fig:ablation}
\end{figure}

We perform an ablation study to showcase the importance of the OC baseline by comparing the success rates of policies learned in the car navigation task with and without it. We choose PPO as the base policy gradient method and augment it with the OC baseline. By default, PPO uses an on-policy value function $V^{\pi}$ as its baseline.
In Fig.~\ref{fig:ablation}, with OC baseline, the robot continuously improves its policy and finally acquires a collision-free, goal-reaching policy with a success rate of around 100\%. In contrast, the standard PPO, which is used without the OC baseline, struggles to develop effective behaviours. This indicates that the RL trial-and-error exploration is data-inefficient and the policy can get stuck to local optima when reward is sparse, but the OC baseline utilizes system and task priors from the associated optimal control problem to improve policy learning.

From another perspective, it is worth noting that the control solution obtained from the associated optimal control problem is usually not directly applicable to the original RL problem. This gap arises because the chosen robot system model is typically an abstraction and simplification of the actual, unknown robot model, plus the optimal control input space may differ from the RL action space.

\subsection{Applied OC baseline to Quadrotor Task}

\begin{figure}[htbp]
    \centering
	\begin{minipage}{\linewidth}
     \centering
 \includegraphics[width=0.58\textwidth]{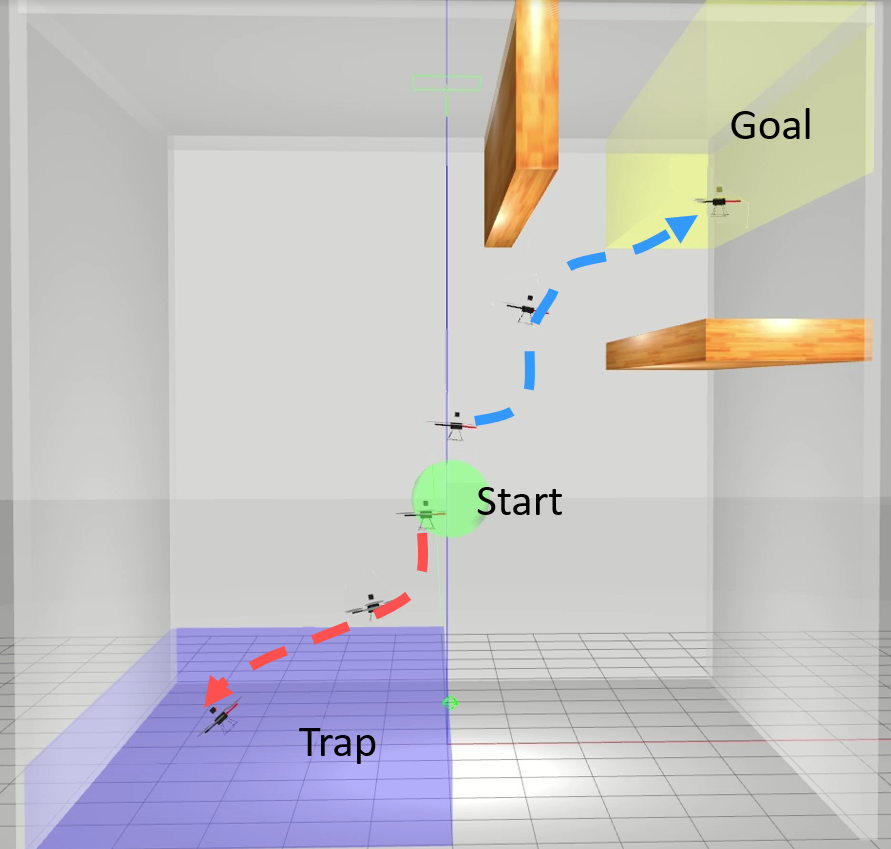}
	\end{minipage}
        \begin{minipage}{\linewidth}
        \centering
        \includegraphics[width=\linewidth]{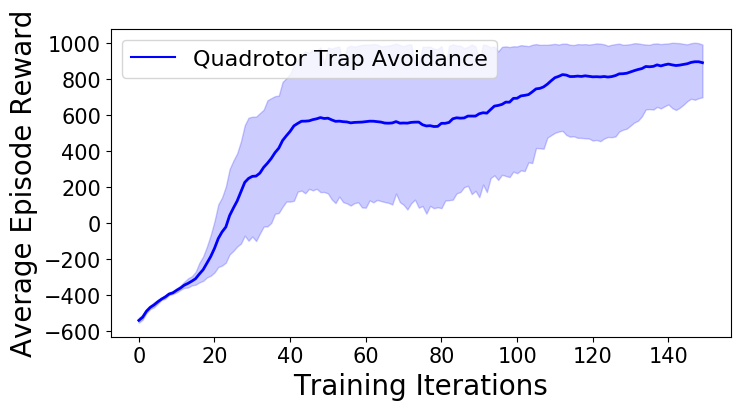}
        \caption{Quadrotor trap avoidance task environment and reward performance}
        \label{fig:quad_nav}
        \end{minipage}
\end{figure}

This method has also been extended to a more complex Quadrotor trap avoidance task, as shown in Fig.~\ref{fig:quad_nav}. We use a detailed planar quadrotor simulated by the Robot Operating System (ROS) \cite{Quigley09} as a complex robot. Controlling a quadrotor is notoriously difficult due to its under-actuated nature. This experiment aims to confirm that our method remains effective even when the robot learning task involves a highly dynamic and unstable system. 

The quadrotor receives RL full state $s=(x, v_x, z, v_z, \psi, \omega, d_1, ..., d_8)$, where $x, z, \psi$ are the planar positional coordinates and pitch angle, and $v_x, v_z, \omega$ denote their time derivatives respectively. It also contains eight sensor readings extracted from the laser rangefinder for obstacle detection. 
$m$ is the  mass of quadrotor, $g$ is gravity, $C^v_{D}$ and $C^\phi_D$ are drag coefficients and $I_{yy}$ is the rigid body inertia based on y-axis. The quadrotor intends to learn a policy mapping from full RL states to its two thrusts $F_1$ and $F_2$ to reach a goal while avoiding the trap. We use a sparse reward function for this task, where $\mathcal{T}$ is the trap area.

\begin{equation}
\small
\label{eq:rew general}
    r(s)=
    \begin{cases}
    0&\;(x,y)\in\mathcal{I}\\
    +1000& \;(x,y)\in\mathcal{G}\\
    -400& \;(x,y)\in\mathcal{C} \\
    +100& \; (x,y)\in\mathcal{T}
    \end{cases}
\end{equation}
 
This task is challenging because, without effective guided exploration, the quadrotor tends to learn a policy that leads it to the trap instead of the goal, due to the positive trap reward and nearby position, which is easily reachable by gravity.

To compute the OC baseline, we choose a 6D planar model \cite{lyu2020ttr} as the simplified abstraction of the fully planar quadrotor, shown in Eq.~\eqref{eq:planar quadrotor main}. The corresponding MPC problem for this task is similar to Eq.~\eqref{eq: mpc for car}, but adding an extra term $-(p_k-\hat{p}^{*})^{\top} \gamma^{k+1} W (p_k-\hat{p}^{*})$ as the penalty of trap-reaching to the cost function, where $\hat{p}^{*}$ is the center position of trap and $W$ is penalty coefficients.
\begin{equation}
\small
\label{eq:planar quadrotor main}
    \dot{\mathrm{x}} =
	\begin{bmatrix}
	\dot x\\
	\dot v_x\\
	\dot z\\
	\dot v_z\\
	\dot \psi\\
	\dot \omega
	\end{bmatrix}
	=
    \begin{bmatrix}
	v_x\\
	-\frac{1}{m}C^v_D v_x + \frac{F_1}{ m}\sin\psi + \frac{F_2}{m}\sin\psi \\
	v_z\\
	-\frac{1}{m}\left(mg+C^v_D v_z\right) + \frac{F_1}{ m}\cos\psi + \frac{F_2}{m}\cos\psi\\
	\omega\\
	-\frac{1}{ I_{yy}}C^\psi_D\omega + \frac{ l}{I_{yy}} F_1 - \frac{l}{I_{yy}} F_2
    \end{bmatrix}
    \hspace{-.4em}
\end{equation}

Fig. ~\ref{fig:quad_nav} depicts the environment and quadrotor's trajectories by executing policies learned with (blue dashed line) and without (red dashed line) our method. We use PPO as the base policy gradient method. With OC baseline, the robot learns the desired goal-reaching policy otherwise can easily lead to a trap-reaching policy without our baseline.

The training results of this experiment are demonstrated in Fig.~\ref{fig:quad_nav}, indicating that in the quadrotor trap avoidance task, the final cumulative reward stabilizes at around 900. Table.~\ref{tab:quad_compar} shows our method has an increasing goal-reaching rate as learning progresses while the normal RL without our baseline has an increasing trap-reaching rate. This result indicates that our method indeed provides guided exploration for the RL process. In this “Trap-Goal” example, although the “Trap” provides a positive reward which should be a good incentive for RL, The OC baseline perceives that the “Goal” region provides even better rewards so that the policy is optimized toward a global optimal direction towards the goal.

\begin{table}[htbp]
\centering
\small
\begin{tabular}{c|c|c}
\hline
Method & Succ rate $\uparrow$ & Trap rate $\downarrow$ \\ \hline
PPO w/o OC baseline&  0.13& 0.65\\ \hline
PPO w/ OC baseline (ours) & \textbf{0.87}& \textbf{0.02} \\ \hline
\end{tabular}%
\caption{\small Comparison of the success rate and trap-reaching rate of Quadrotor trap avoidance task.   }
\label{tab:quad_compar}
\end{table}

\section{Conclusion and future work}
We introduce a novel baseline function for policy gradient RL. Our baseline is derived from the optimal control value function, which is computed based on an optimal control problem that is formed to be closely related to the target RL task. We demonstrate that our baseline is able to ensure guided exploration to improve policy gradient RL especially when the task has sparse and insufficient feedback.

\bibliography{IEEEabrv.bib}
\bibliographystyle{IEEEtran.bst}

\end{document}